\documentclass[10pt,twocolumn,letterpaper]{article}
\usepackage{multirow}
\usepackage{wacv}              

\newcommand{\ProposedBenchmark}{LASER-bench}%

\usepackage{multirow}
\usepackage{amssymb}
\usepackage{pifont}
\usepackage{adjustbox}
\newcommand{\cmark}{\ding{51}}
\newcommand{\xmark}{\ding{55}}

\definecolor{wacvblue}{rgb}{0.21,0.49,0.74}
\usepackage[pagebackref,breaklinks,colorlinks,allcolors=wacvblue]{hyperref}

\def\wacvPaperID{285} 
\def\confName{WACV}
\def\confYear{2026}

\title{LASER: Lip Landmark Assisted Speaker Detection for Robustness}
\begin{document}
\author{Le Thien Phuc Nguyen*
\qquad
Zhuoran Yu\thanks{equal contribution}
\qquad
Yong Jae Lee
\\
University of Wisconsin - Madison\\
{\tt\small \{plnguyen6, zhuoran.yu\}@wisc.edu}\\
{\tt\small yongjaelee@cs.wisc.edu} \\
\small \url{https://github.com/plnguyen2908/LASER_ASD}\\
\small \url{https://huggingface.co/datasets/plnguyen2908/LASER-bench}
}

\newcommand{\ProposedMethodName}{{{{LASER}}}}

\maketitle

\begin{abstract}
    Active Speaker Detection (ASD) aims to identify who is speaking in complex visual scenes. While humans naturally rely on lip-audio synchronization, existing ASD models often misclassify non-speaking instances when lip movements and audio are unsynchronized. To address this, we propose \textbf{L}ip landmark \textbf{A}ssisted \textbf{S}peaker d\textbf{E}tection for \textbf{R}obustness (\ProposedMethodName{}), which explicitly incorporates lip landmarks during training to guide the model’s attention to speech-relevant regions. Given a face track, \ProposedMethodName{} extracts visual features and encodes 2D lip landmarks into dense maps. To handle failure cases such as low resolution or occlusion, we introduce an auxiliary consistency loss that aligns lip-aware and face-only predictions, removing the need for landmark detectors at test time. \ProposedMethodName{} outperforms state-of-the-art models across both in-domain and out-of-domain benchmarks. To further evaluate robustness in realistic conditions, we introduce \ProposedBenchmark{}, a curated dataset of modern video clips with varying levels of background noise. On the high-noise subset, \ProposedMethodName{} improves mAP by 3.3 and 4.3 points over LoCoNet and TalkNet, respectively, demonstrating strong resilience to real-world acoustic challenges.


\end{abstract}

    
\section{Introduction}
\label{sections:intro}

Active Speaker Detection (ASD)~\cite{jung2024talknce, wang2024loconet, tao2021someone, liao2023light, roy2021learning, alcazar2021maas} is a fundamental task in audiovisual computing which aims to detect if one or more people are speaking in a complex visual scene (usually represented by videos). To effectively accomplish this task, models are not only required to understand the visual scene, extract descriptive visual features of candidate speakers, but also learn to correspond the visual information to corresponding audio information. Advanced applications in audiovisual computing such as human-robot interaction \cite{kang2023video, sheridan2016human, skantze2021turn} and multimodal chatbots \cite{wu2023next, shu2023audio, ge2024worldgpt} usually rely on the success of ASD.

\begin{figure}[t]
  \centering
  \begin{minipage}{\columnwidth}
  \centering
    \begin{subfigure}{\columnwidth}
      \includegraphics[width=\linewidth]{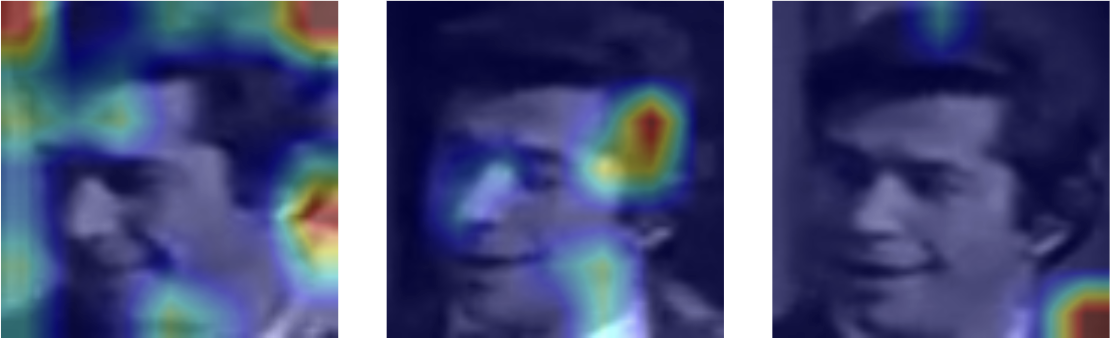}
      \caption{Grad-CAM for LoCoNet.}
      \label{fig:vis-baseline}
    \end{subfigure}
    \hfill
    \begin{subfigure}{\columnwidth}
      \includegraphics[width=\linewidth]{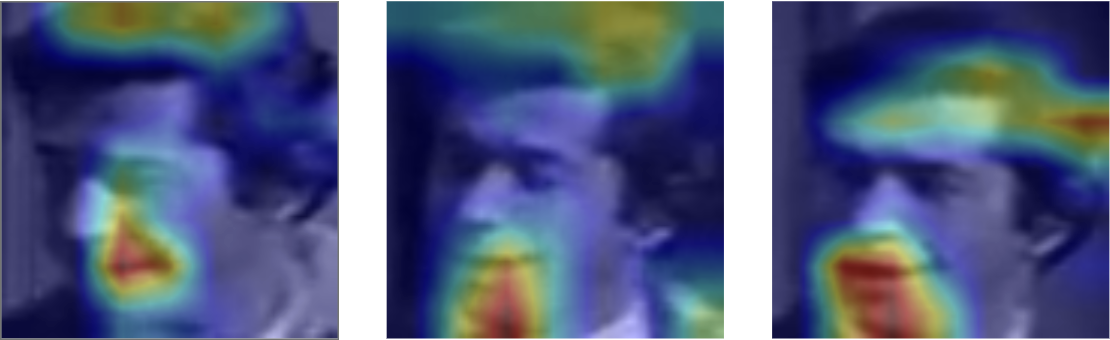}
      \caption{Grad-CAM for our \ProposedMethodName{}}
      \label{fig:vis-ours}
    \end{subfigure}
  \end{minipage}

\vspace{-2mm}
\caption{\textbf{Qualitative comparison of Grad-Cam~\cite{grad-cam} visualizations for LoCoNet and our \ProposedMethodName{} under nonsynchronized audiovisual scenarios}. LoCoNet~\cite{wang2024loconet} struggles to accurately predict ``not speaking'' when visual frames and audio tracks are misaligned, often failing to focus on the lip region when making these predictions. In contrast, our \ProposedMethodName{} consistently concentrates on the lip area and successfully identifies ``not speaking'' situations.}
\vspace{-10pt}
\label{fig:visualization}
\end{figure}


How can we, as humans, tell whether a person is speaking? The synchronization of mouth movements with speech is a natural indicator, allowing us to effortlessly perceive when someone is actually speaking~\cite{Park2016LipME}. Even a slight delay that desynchronizes audio and visual cues creates discomfort, making it immediately apparent that the person is not truly speaking at that moment. This inherent sensitivity to audiovisual alignment highlights the central role of mouth movement in active speaker detection.



Do existing ASD models function in the same intuitive way, and can they detect when a person is not speaking if the audio and mouth movements are unsynchronized? Unfortunately, based on our analysis, the answer to both questions is no. State-of-the-art ASD models consist of two main phases: (1) visual and audio temporal representation learning, which encodes video and audio streams into sequences of visual and audio embeddings with temporal context, and (2) inter- and intra-person context modeling, which leverages cross-attention to capture long-term temporal dependencies and visual-audio interactions. 
While current approaches utilize attention mechanisms to model inter- and intra-person context \cite{wang2024loconet, datta2022asd, alcazar2020active, roy2021learning, alcazar2021maas}, they still struggle to maintain a focus on mouth movement when making prediction (Figure~\ref{fig:vis-baseline}). This limitation prevents models from reliably identifying unsynchronization between audio and visual cues, a key factor in human perception.

While most existing work focuses on improving context modeling, we argue that such limitations stem from the visual features themselves, even before they reach the context modeling module. To address this, we propose \textbf{L}ip landmark \textbf{A}ssisted \textbf{S}peaker d\textbf{E}tection for \textbf{R}obustness (\ProposedMethodName{}) to overcome these limitations. The key novelty of \ProposedMethodName{} lies in the lightweight visual temporal representation learning of ASD models: instead of relying on the model to learn audiovisual interactions purely from facial frames, \ProposedMethodName{} explicitly directs the model’s attention to lip movements by incorporating lip landmarks in training (Figure~\ref{fig:vis-ours}).

Given a face track\footnote{A face track is a sequence of cropped images of the same person's face across consecutive frames in a video stream~\cite{roth2020ava}.}, \ProposedMethodName{} extracts frame-level visual feature maps and the 2D coordinates of lips landmarks for each frame of the face track using a lightweight facial landmark detector~\cite{lugaresi2019mediapipe}, forming a ``lip track'' corresponding it. We then encode the landmark coordinates to continuous 2D feature maps matching the spatial dimension of the visual feature maps and combine them to form our visual features that combine the information from the face track and the lip track.  Since the lip landmarks come in as 2D coordinates instead of continuous visual feature maps, integrating such discrete coordinates with continuous feature maps poses a challenge. To address this, we design a simple-yet-effective encoding strategy that converts the 2D discrete coordinate information into continuous 2D visual feature maps matching the spatial dimensions of the corresponding visual features (details in Section~\ref{sections:mler}). Furthermore, to alleviate the sparsity of these feature maps, we aggregate them into dense feature maps through a 1x1 convolution layer. This way, the lip landmark coordinates are converted to dense feature maps carrying out the information on the position and form of the lip.  The combined visual features are passed through the visual temporal network to obtain the final visual representations with temporal context and the representation can then be used by selected ASD models to capture long-term temporal dependencies and integrate the information from multimodal features.



While \ProposedMethodName{} enhances ASD model performance, we observe that facial landmark detectors may occasionally fail to provide reliable outputs due to issues like low resolution, occlusion, or extreme facial angles. For example, on widely used AVA benchmarks~\cite{roth2020ava}, approximately 15\% of face tracks in the test sets lack any landmark detection results. Consequently, missing lip landmarks—resulting in the absence of lip track information—can lead to suboptimal model performance on these test videos. To address this challenge, we introduce an auxiliary consistency loss that aligns predictions made using \ProposedMethodName{} with those based solely on visual features from the face track. Specifically, for each frame of a face track, we optimize a KL-divergence loss between the predictions obtained with \ProposedMethodName{} and those based on visual face-frame features alone. This auxiliary loss enables the model to maintain accurate predictions even when the lip track is unavailable, eliminating the dependency on a facial landmark detector at test time.

Since \ProposedMethodName{} leverages explicit lip information, it is particularly suited for acoustically noisy environments where accurate lip-audio correspondence is essential. To evaluate this, we introduce \ProposedBenchmark{}, a curated benchmark featuring modern video content with authentic background noise, including crowd chatter and ambient sounds. Each clip is processed using a voice activity detection (VAD) tool~\cite{SileroVAD} to isolate non-speech segments, and categorized into low and high-noise subsets based on RMS energy. This allows for controlled analysis of model robustness under varying noise levels in realistic settings.

To evaluate the effectiveness of our approach, we adopt multiple protocols. Under the standard setup, we train on AVA~\cite{roth2020ava} and evaluate on both in-domain (AVA) and out-of-domain (Talkies~\cite{alcazar2021maas}, ASW~\cite{kim2021look}) test sets, where \ProposedMethodName{} consistently outperforms LoCoNet. To assess noise robustness, we evaluate on \ProposedBenchmark{}, which contains modern video clips grouped into low and high-noise subsets using VAD-filtered RMS thresholds. \ProposedMethodName{} achieves notable gains over LoCoNet~\cite{wang2024loconet} and TalkNet~\cite{tao2021someone}, particularly under high-noise conditions, up to +3.3 and +4.3 mAP, respectively. We also test robustness to audio-visual misalignment using audio-swapped and delayed audio settings, where our method maintains consistent improvements in frame-level accuracy across all datasets. 

\vspace{-10pt}
\paragraph{Contributions.} 1) We propose \ProposedMethodName{}, a novel ASD approach that explicitly directs the model's attention to lip movements by incorporating lip landmarks in training through effective landmark encoding. 2) We introduce an auxiliary consistency loss during training to address the absence of lip track information at test time, eliminating dependency on facial landmark detectors during inference. 3) We create \ProposedBenchmark{}, a curated benchmark designed to systematically evaluate ASD robustness under real-world acoustic conditions. 4) \ProposedMethodName{} outperforms state-of-the-art models under standard evaluation protocols, and further shows improved robustness in real-world scenarios such as audio-visual misalignment and high background noise.
\section{Related Work}
\label{sections:related_works}

\paragraph{Active Speaker Detection.} Active Speaker Detection (ASD) is a key audiovisual challenge, where the goal is to determine if someone is speaking in a video. Recent work~\cite{tao2021someone, wang2024loconet, jung2024talknce, alcazar2021maas, roy2021learning, liao2023light} emphasizes long-term temporal modeling via RNNs~\cite{liao2023light, kopuklu2021design}, attention~\cite{wang2024loconet, datta2022asd, ADENet}, or both~\cite{alcazar2020active}. LoCoNet~\cite{wang2024loconet} adopts self- and cross-attention for intra-speaker context, plus CNNs for inter-speaker modeling; TalkNCE~\cite{jung2024talknce} extends LoCoNet with contrastive learning. In contrast, our method shifts the focus by explicitly highlighting lip movements, complementing these context-driven approaches. As shown in our experiments, augmenting existing ASD models with \ProposedMethodName{} yields stronger generalization performance under high background noise context and other complex scenarios.

\vspace{-11pt}
\paragraph{Facial Landmark Guidance in Audiovisual Computing.}
Earlier studies (Saenko et al.~\cite{saenko2005visual}, Everingham et al.~\cite{everingham2009taking}) extracted facial features by localizing key regions. For example, Saenko et al.~\cite{saenko2005visual} focused on lip localization to detect speaking activity. Recent works have explored the integration of facial landmarks to improve model performance on audiovisual computing tasks by using facial landmarks for feature pooling~\cite{Wang2019landmark_pooling} or directly incorporating discrete landmark coordinates~\cite{Geeroms2022_2d_encoding}. While these prior approaches align with our intuition that speaking behavior is intricately linked to lip movements, they fundamentally differ from our method. Our approach transforms the 2D coordinates of lip landmarks into sparse feature maps through a simple-yet-effective encoding function and our approach outperforms these alternative designs within the state-of-the-art approach LoCoNet framework (Section~\ref{section:alternative}).

\vspace{5pt}
\noindent \textbf{Active Speaker Detection Datasets.} The field of active speaker detection has been significantly advanced by several key benchmarking datasets. AVA-ActivateSpeaker~\cite{roth2020ava} stands out as the largest dataset in this domain, featuring videos derived from movies over a decade ago. In contrast, Talkies~\cite{alcazar2021maas} and ASW~\cite{kim2021look} curate data from YouTube and the VoxConverse dataset~\cite{voxConverse}, respectively, focusing on real-world scenarios to provide more diverse speaking environments. Since our approach explicitly incorporates lip information, it is conceptually better equipped to handle acoustically noisy conditions, where strong lip-audio correspondence becomes crucial. To evaluate this, we introduce \ProposedBenchmark{}, a benchmark spanning a wide range of real-world speaking scenarios under background noise for comprehensive assessment.
\section{Approach}
\label{sections:method}
In this section, we present the details of our proposed \textbf{L}ip landmark \textbf{A}ssisted \textbf{S}peaker d\textbf{E}tection for \textbf{R}obustness (\ProposedMethodName{}). \ProposedMethodName{} is a versatile training framework that can be integrated with state-of-the-art ASD models. We begin by reviewing the general framework of ASD models (Section ~\ref{sections:preliminary}), then describe how we incorporate lip movement guidance into the model (Section~\ref{sections:mler}), and finally introduce our consistency loss, which addresses the issue of missing lip landmarks at test time (Section~\ref{sections:consistency}).

\subsection{Preliminaries}
\label{sections:preliminary}

The goal of active speaker detection is to classify the speaking activity 
\(Y \in \mathbb{R}^{T}\) for every frame of a face track \(V \in \mathbb{R}^{T \times H \times W}\) conditioned on audio Mel-spectrograms \(A \in \mathbb{R}^{4T \times N}\). Here, $T$ is face track's temporal length, $(H, W)$ is the face crop's spatial dimension, and $N$ is the number of audio Mel-spectrogram frequency bins.

State-of-the-art ASD models~\cite{wang2024loconet, tao2021someone, liao2023light, datta2022asd} typically consist of two main components: audio-visual encoders and long-term intra-speaker context modeling modules. The audio-visual encoder includes separate visual and audio encoders, \(\mathcal{F}_v\) and \(\mathcal{F}_a\), respectively. The visual encoder \(\mathcal{F}_v\) processes visual face tracks to produce visual features \(f_v \in \mathbb{R}^{T \times D}\), while the audio encoder \(\mathcal{F}_a\) encodes audio spectrograms, generating audio features \(f_a \in \mathbb{R}^{T \times C}\), where \(D\) and \(C\) denote the embedding dimensions for visual and audio features, respectively.  $f_v$ and $f_a$ are then concatenated to $f_{av} = concat(f_v, f_a)$ to form multimodal features and $f_{av}$ is further processed by the context modeling modules $\mathcal{G}$~\cite{wang2024loconet, tao2021someone, liao2023light, datta2022asd} to aggregate information from both modalities. Typically, $\mathcal{G}$ outputs multimodal features $f'_{av}$ and unimodal features $f'_v$ and $f'_a$. Linear classifiers, one for each of the three modalities, are then used to process the corresponding features to make final predictions. $\mathcal{F}$, $\mathcal{G}$ and the linear classifiers are trained end-to-end with the following objective:
\begin{equation}
\label{eq:asd}
 \mathcal{L_\text{asd}} = \lambda_v \cdot \mathcal{L}_v + \lambda_a \cdot \mathcal{L}_a + \lambda_{av} \cdot \mathcal{L}_{av}
\end{equation}
where $\mathcal{L}_{av}$ is the cross-entropy loss computed between ground-truth labels $GT$ and predictions $Y$ in each frame and $L_a$ and $L_v$ are auxiliary cross-entropy losses computed with ground-truth labels and predictions using the unimodal features $f'_v$ and $f'_a$, respectively, to avoid $\mathcal{G}$ overly focusing on one modality of features~\cite{roth2020ava}. 

\begin{figure*}[t]
  \centering
    \includegraphics[width=\linewidth]{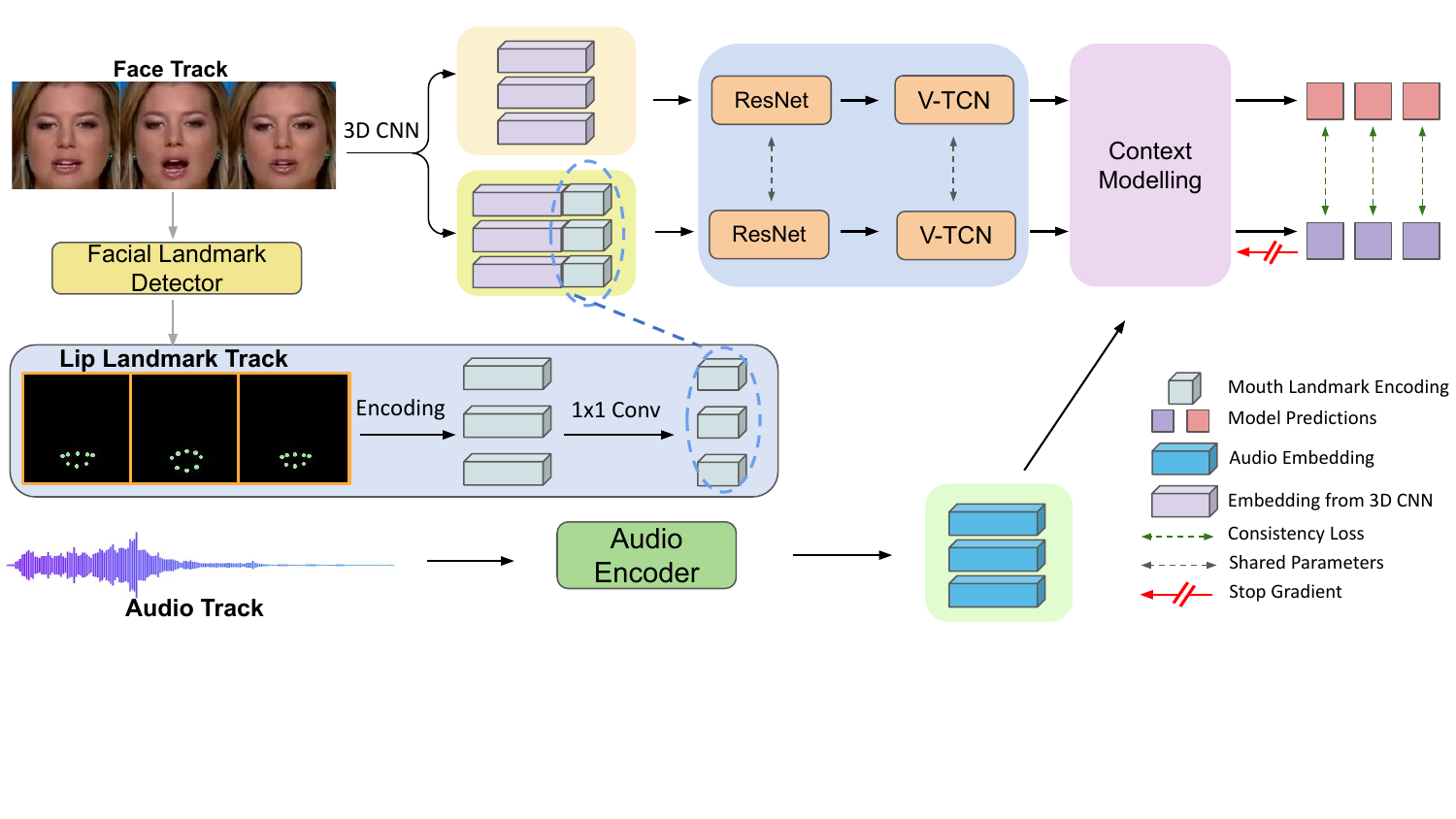}
  \vspace{-3cm}
  \caption{\textbf{Illustration of \ProposedMethodName{}}. Given a face track \(V\), we first obtain a lip landmark track using a facial landmark detector and encode the 2D coordinates of these landmarks into continuous 2D feature maps. These maps are then aggregated through a 1x1 convolution layer. The encoded lip track is concatenated with visual features from a 3D CNN and fed into ResNet and V-TCN to capture a temporal visual representation, which is further processed by context modeling modules~\cite{wang2024loconet, tao2021someone, liao2023light} to produce the final prediction. The consistency loss is computed between predictions made with and without lip landmark encoding and gradients are only propagated through the prediction without lip landmark. This way, the model is robust against missing lip landmarks at test time.}
  \vspace{-5mm}
  \label{fig:pipeline}
\end{figure*}

\subsection{Enhanced Representation from Lip Landmark Guidance}
\label{sections:mler}

While existing work~\cite{wang2024loconet, tao2021someone, liao2023light, datta2022asd} demonstrates competitive performance on the popular AVA-ActiveSpeaker Benchmark~\cite{roth2020ava}, our analysis reveals that they still struggles to capture the synchronization between mouth movements and the audio track (Figure~\ref{fig:vis-baseline}). This limitation highlights the need for enhanced focus on lip-audio synchronization within active speaker detection models.

We speculate that this limitation stems from the visual features themselves, even before they reach the context modeling module. To address this, we propose \ProposedMethodName{}, which focuses on enhancing the representations produced by the visual encoder \(\mathcal{F}_v\). By improving these initial visual features, the context modeling module \(\mathcal{G}\) can more effectively integrate visual and audio information, ultimately enhancing speaker detection performance.

We adopt the same visual encoder architecture $\mathcal{F}_v$ from prior work~\cite{tao2021someone, wang2024loconet} where the architecture is composed on a 3D convolution layer, a ResNet-18~\cite{he2016deep} backbone, and a video temporal network (V-TCN)~\cite{lea2016temporal}. However, unlike prior work which simply forwards visual face tracks to $\mathcal{F}_v$, we additionally apply a facial landmark detector to the face track and collect a ``lip landmark track'' \(M\) of the corresponding face track. Each lip landmark is represented as 2D coordinates ${(x_i, y_i)}_{i=0}^K$ where $K$ is the total number of lip landmarks collected. 

\subsubsection{Lip Landmark Track Encoding}
\label{section:lte}
Integrating discrete 2D coordinate information with continuous visual signals during training poses a challenge. To address this, we design a simple-yet-effective encoding function \(\mathcal{T}(M)\) that converts the 2D discrete coordinate to continuous 2D feature maps matching the spatial dimensions of the visual features. Specifically, the encoding process \(h = \mathcal{T}(M)\) for each landmark ${(x_i, y_i)}$ is defined as follows:
\begin{equation}
    h_x^i (u, v) = \begin{cases} 
x_i / W, & \text{if } u = x_i \,,  \, v = y_i \\ 
0, & \text{if } \text{otherwise} 
\end{cases}
\end{equation}

\begin{equation}
    h_y^i (u, v) = \begin{cases} 
y_i / H, & \text{if } u = x_i \,, \, v = y_i \\ 
0, & \text{if } \text{otherwise} 
\end{cases}
\end{equation}

The encoding process transforms the ``lip landmark track'' into features \( h \in \mathbb{R}^{T \times K \times 2 \times H \times W} \), where \( K \) is the number of lip landmarks, and 2 corresponds to the \( x \) and \( y \) coordinates. We use the lightweight facial landmark detector from Mediapipe~\cite{lugaresi2019mediapipe} to extract \( K = 82 \) lip landmarks.

\subsubsection{Lip Landmark Track Feature Aggregation}

The encoded landmark features are sparse and contain many redundant zero entries, which can hinder optimization during training. To address this, we apply a simple feature aggregation module using $1 \times 1$ convolution layers that condense the information from \(K\) channels down to \(S\) channels, where \(S \ll K\). In practice, we find that setting \(S = 4\) achieves the best performance, adding only negligible computational overhead. We denote the aggregated landmark feature as \(h' \in \mathbb{R}^{T \times S \times 2 \times H \times W} \).

We concatenate the lip landmark track feature \( h' \) with face track visual features from a 3D convolution layer, passing the combined feature through a ResNet for frame-level processing. The output is then fed into a video temporal network~\cite{lea2016temporal} to generate temporal visual representations \( f_{v*} \in \mathbb{R}^{T \times D} \). In this way, the visual encoder not only incorporates cues from frame-level lip landmarks but also leverages temporal modeling to explicitly capture the dynamics of lip movement across video frames. 

Our visual encoder can be integrated to state-of-the-art ASD models~\cite{wang2024loconet, tao2021someone, liao2023light, datta2022asd} and can be trained end-to-end by the ASD training objectives described in Equation~\ref{eq:asd}. As we shall see in the experiments, \ProposedMethodName{} enhances state-of-the-art ASD models across various evaluation protocols.

\subsection{Addressing Missing Lip Landmark Tracks}
\label{sections:consistency}

While \ProposedMethodName{} achieves strong performance across various evaluation scenarios, we observe that the facial landmark detector does not always return lip landmarks for every face track because of low resolution, occlusion, or extreme angles in the facial frames. For example, our facial landmark detector fails to detect any landmarks on approximately 15\% of the face track on the AVA-ActiveSpeaker dataset~\cite{roth2020ava}. At test time, the absence of lip landmarks results in suboptimal visual features, which can negatively impact the model's performance.

To enable our visual encoder \(\mathcal{F}_v\) to effectively handle face tracks that are challenging for landmark detectors, we introduce an auxiliary consistency loss to enforce consistency between the predictions made from \ProposedMethodName{}-enhanced features and those from standard face track features. Specifically, given a face track \( V \in \mathbb{R}^{T \times H \times W} \), we obtain the visual representation \( f_v = \mathcal{F}_v(V) \) without lip track encoding and \( f_{v*} = \mathcal{F}_v(V, \mathcal{T}(M)) \) with lip track encoding, where \(\mathcal{T}(M)\) represents the lip landmark encoding. The consistency loss is then defined as the KL-Divergence between the final predictions made with \( f_v \) and \( f_{v*} \): 
\begin{equation}
\mathcal{L}_{\text{consistency}} = \sum_{j=0}^1 \mathcal{G}(f_v)_j \cdot \log \frac{\mathcal{G}(f_v)_j}{\mathcal{G}(f_{v*})_j}
\end{equation}
where \(j\) denotes the class index of prediction. We apply stop gradient operation on $f_{v*}$ (predictions made with lips track encoding) when computing the consistency loss so that gradients only flow through $f_v$ (predictions without lips track encoding), encouraging the model to match the predictions it makes with lips track encoding, implicitly guiding the model to focus on the lips-audio synchronization even when the lips landmarks are not provided. 

The final training objective is defined as
\begin{equation}
 \mathcal{L} = \mathcal{L}_{\text{asd}} + \lambda_c \mathcal{L}_{\text{consistency}}
  \label{final_loss}
\end{equation}
where \(\mathcal{L}_{\text{asd}}\) is the default training objective of ASD models~\cite{wang2024loconet, tao2021someone, liao2023light, datta2022asd} from Section~\ref{sections:preliminary}.  By enforcing consistent predictions between face tracks with and without lip track encoding, our model achieves robust performance even on challenging face tracks where facial landmark detection may fail. Notably, as demonstrated in our experiments, this approach allows us to completely eliminate the reliance on the facial landmark detector, using only face tracks at test time without sacrificing performance.

\section{\ProposedBenchmark{}}

Recall that \ProposedMethodName{} leverages explicit lip information to improve audio-visual correspondence, which is particularly valuable in acoustically noisy settings where background sounds may interfere with speech perception. However, existing ASD benchmarks offer limited support for evaluating this aspect. AVA-ActiveSpeaker~\cite{roth2020ava}, though large in scale, consists mainly of clean audio from scripted movie scenes, while ASW~\cite{kim2021look} includes noisy, in-the-wild videos but remains small in size. To bridge this gap, we introduce \ProposedBenchmark{}, a large-scale benchmark designed to evaluate ASD robustness under authentic background noise.

\begin{table*}[t]
    \centering
    
    \begin{tabular}{l|ccc|c}
        \toprule
        \textbf{Statistics} & \textbf{AVA-Val \cite{roth2020ava}} & \textbf{Talkies-Val \cite{alcazar2021maas}} & \textbf{ASW-Val \cite{kim2021look}} & \textbf{\ProposedBenchmark{} (Ours)}  \\
        \midrule
        Total face tracks          & 8K    & 6.8K    & 3.5K      & 4.9K \\
        Total face crops           & 760K  & 235K   & 171K & 738K   \\
        \bottomrule
    \end{tabular}
    \vspace{-3mm}
    \caption{\textbf{Dataset comparison.} We compare relevant statistics of MADE with three other well-known datasets for active speaker detection (left columns) and \ProposedBenchmark{} (right columns).}
    \vspace{-3mm}
    \label{tab:dataset}
\end{table*}

\subsection{Data Curation.}
In creating \ProposedBenchmark{}, we aim to assemble a diverse dataset that captures complex acoustic scenarios, including off-screen speech, crowd noise, and environmental noise. To achieve this goal, we primarily use YouTube to source our clips, focusing on modern videos that typically include spoken dialogue and multiple visible individuals. After gathering an initial pool of \textit{vlogs}, \textit{street interviews}, \textit{podcasts/interviews}, and \textit{multi-person TV shows} through relevant search terms, we then filter out any content involving sensitive or potentially objectionable topics. Then, we follow the procedure described in AVA-ActiveSpeaker~\cite{roth2020ava} by keeping the face tracks that are at least 0.2 seconds and interpolating to create missing detection inside the tracks. However, we do not limit the length of each track like AVA-ActiveSpeaker~\cite{roth2020ava}, and every annotation is gone through two rounds of verification by our annotators.

\subsection{Dataset Statistics}

Table \ref{tab:dataset} provides a concise comparison of four active‐speaker benchmarks, listing total face tracks and face crops for each: AVA‐Val~\cite{roth2020ava} (8K tracks, 760K crops), Talkies‐Val~\cite{alcazar2021maas} (6.8K tracks, 235K crops), ASW‐Val~\cite{kim2021look} (3.5K tracks, 171K crops) and our \ProposedBenchmark{} (4.9K tracks, 738K crops). Notably, while ASW‐Val incorporates real background noise to simulate challenging acoustic conditions, LASER‐bench preserves that same noise realism at a much larger scale - offering over 1.4× more face tracks and more than 4× the number of face crops - thereby furnishing a richer resource for evaluating speaker detection models under noisy, in‐the‐wild scenarios.

\section{Experiments}

In this section, we evaluate the effectiveness of our proposed approach (\ProposedMethodName{}). We assess its performance on multiple datasets and compare it against baseline models to demonstrate improvements across different evaluation scenarios. We also ablate our design choices.

\subsection{Experimental Setup}
\label{sections:experimental_setup}

\subsubsection{Implementation Details} 
 We implement \ProposedMethodName{} with the ASD models including LoCoNet~\cite{wang2024loconet}, TalkNet~\cite{tao2021someone}, and Light-ASD~\cite{liao2023light}. We closely follow the training details of each model for fair comparison (details in Appendix). For ASD training objectives, we follow the LoCoNet and TalkNet, setting $\lambda_{av} = 1$, $\lambda_a = 0.4$, and $\lambda_v = 0.4$. Random resizing, cropping, horizontal flipping, and rotations are used as visual data augmentation operations and a randomly selected audio signal from the training set is added as background noise to the target audio~\cite{tao2021someone}. For LASER, We set $S = 4$ and $\lambda_c = 1$ for all evaluations and ablation studies of these hyper-parameters can be found in Appendix.


\subsubsection{Evaluation Protocols} We use the following evaluation protocols to assess the performance of \ProposedMethodName{} and baseline methods:

(1) \textbf{Standard Evaluation}: We evaluate on both in-domain scenarios (where training and testing data come from the same dataset) and out-of-domain scenarios (where testing data comes from different datasets). We use AVA-ActiveSpeaker~\cite{roth2020ava} as our in-domain datasets and Talkies~\cite{alcazar2021maas} and ASW~\cite{kim2021look} as our out-of-domain evaluation datasets and use mAP as the evaluation metric following common practice~\cite{roth2020ava}.

(2) \textbf{Evaluation on \ProposedBenchmark{}}. In addition to the synchronization schema, we evaluate \ProposedMethodName{} on \ProposedBenchmark{} using the same metric mentioned above. In particular, we divide \ProposedBenchmark{} into two subsets: Low Noise and High Noise. To do so, for each audio track, we first remove speech segments using a Voice Activity Detection (VAD) tool~\cite{SileroVAD}, then compute the Root Mean Square (RMS) energy over the remaining background audio. A threshold of 0.03 RMS distinguishes low and high noise levels.

(3) \textbf{Evaluation with Unsynchronized audios}: This protocol simulates non-speaking scenarios where visual frames and audio tracks are misaligned, a critical test for applications like web conferencing. For instance, if a muted participant talks to someone else while the main speaker is active, the model should correctly classify the muted participant as ``not speaking'' to avoid confusion. We evaluate using two protocols: 1) replacing the original audio with another video's audio and 2) introducing a short audio delay. The evaluation uses the same dataset in (1), with results reported as average accuracy across all frames. \footnote{mAP is unsuitable here as all ground-truth frames are negative (non-speaking).}

\begin{figure*}[t]
  \centering
  \begin{subfigure}{0.33\linewidth}  
    \includegraphics[width=\linewidth]{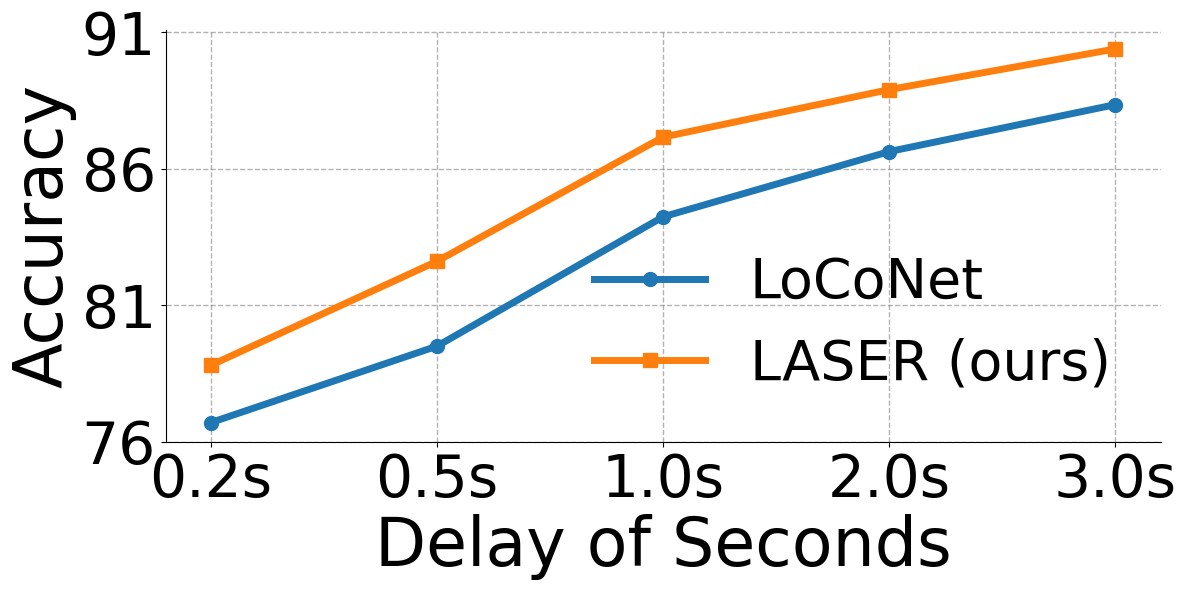}  
    \caption{Evaluation on AVA-ActiveSpeaker}
  \end{subfigure}  
  \begin{subfigure}{0.33\linewidth}  
    \includegraphics[width=\linewidth]{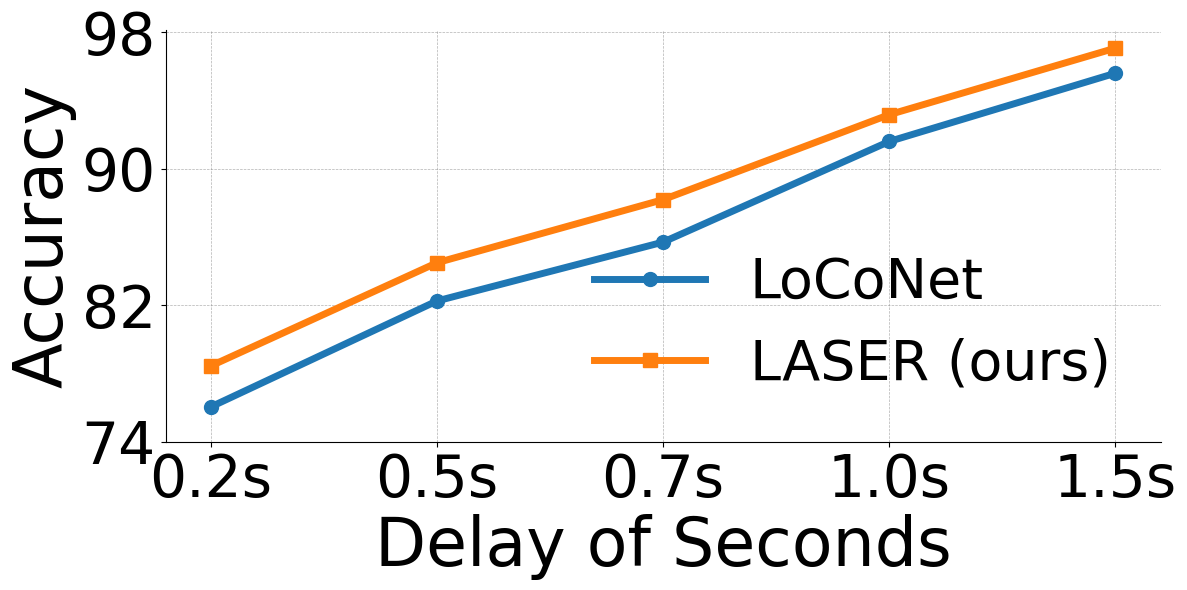}  
    \caption{Evaluation on Talkies}
  \end{subfigure}
  \begin{subfigure}{0.33\linewidth}  
    \includegraphics[width=\linewidth]{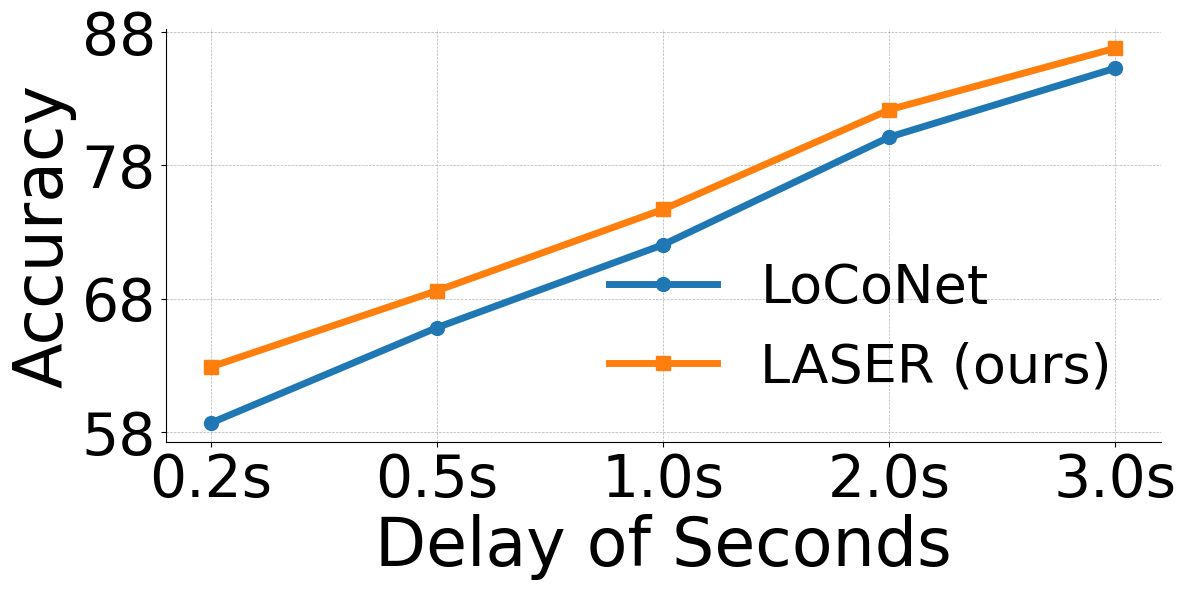}  
    \caption{Evaluation on ASW}
  \end{subfigure}
  \caption{\textbf{Evaluation with unsynchronized audios}. We use the same datasets as the evaluation of synchronized audios and report the per-frame accuracy on in-domain datasets (AVA-ActiveSpeaker) and out-of-domain datasets (Talkies and ASW). \ProposedMethodName{} consistently outperforms LoCoNet~\cite{wang2024loconet} under this evaluation protocol.}
  \label{fig:unsync}
\end{figure*}

\begin{table}[t]
  \centering
  
 \begin{tabular}{lccc}
    \toprule
    \textbf{Model} & \textbf{AVA} & \underline{\textbf{Talkies}} & \underline{\textbf{ASW}}  \\
    \midrule
    EASEE \cite{alcazar2022end} & 94.1 & 86.7 & -\\
    MAAS \cite{alcazar2021maas} & 88.8 & 79.7 & -\\
    TalkNet \textsuperscript{\dag} \cite{tao2021someone} & 92.2  & 85.9 & 85.8\\
    \textbf{TalkNet + \ProposedMethodName{}} & \textbf{92.5} & \textbf{87.6} & \textbf{86.6}\\
    \midrule
    Light-ASD\textsuperscript{\dag} \cite{liao2023light} & 93.5 & 87.6 & 87.4 \\
    \textbf{Light-ASD + \ProposedMethodName{}} & \textbf{93.8} & \textbf{88.1} & \textbf{87.6}\\
    \midrule
    LoCoNet \textsuperscript{\dag} \cite{wang2024loconet} & 95.2  & 88.4 & 88.4 \\
    \textbf{LoCoNet + \ProposedMethodName{}} &  \textbf{95.3} & \textbf{89.0} & \textbf{88.9} \\
    \midrule
    LoCoNet w/ CL \textsuperscript{\dag} \cite{jung2024talknce} & \textbf{95.5} & 88.3 & 88.5  \\
    \textbf{LoCoNet w/ CL + \ProposedMethodName{}} & 95.4 & \textbf{89.7} & \textbf{89.5} \\
    \bottomrule
  \end{tabular}
  \vspace{-1mm}
  \caption{\textbf{Results of models trained on AVA-ActiveSpeaker~\cite{roth2020ava}}. The models are trained on AVA-ActiveSpeaker dataset only and evaluated on AVA, as in-domain evaluation and Talkies and ASW as out-of-domain evaluations (underlined). CL denotes contrastive learning. \textsuperscript{\dag} Results are reproduced by the methods' codebase.}
  \vspace{-5mm}
  \label{tab:sync}
\end{table}

\subsection{Standard Evaluation} 
\label{section:eval_ava}
We further evaluate \ProposedMethodName{} on AVA-ActiveSpeaker, the most widely used benchmark for active speaker detection. Training is conducted on the AVA training set, with in-domain evaluations on the AVA-val set and out-of-domain evaluations on Talkies~\cite{alcazar2021maas} and ASW~\cite{kim2021look}, following LoCoNet~\cite{wang2024loconet}.

As shown in Table~\ref{tab:sync}, \ProposedMethodName{} still improves upon LoCoNet for in-domain evaluations when the performance is already saturated on AVA. When evaluating on Talkies and ASW whose data distribution is different from AVA, \ProposedMethodName{} achieves 89.0 and 88.9 mAP respectively. Moreover, for efficient architectures like TalkNet or Light-ASD, \ProposedMethodName{} advances the performance on all three benchmarks. For example, \ProposedMethodName{} improves TalkNet's evaluation on Talkies and ASW by 1.7 and 0.8 mAP respectively. 

Recent work~\cite{jung2024talknce} also proposes to add an auxiliary contrastive loss to LoCoNet which brings audio-visual embedding pairs closer when they originate from the same video frame, while pushing them apart otherwise. As shown in Table~\ref{tab:sync} (bottom), while the contrastive loss enhances LoCoNet's in-domain performance by 0.3 mAP, it offers limited gains in out-of-domain scenarios. Interestingly, when combined with \ProposedMethodName{}, our method achieves an additional 1.4 and 1.0 mAP improvement on the Talkies and ASW benchmarks, respectively. These results highlight both the effectiveness and general applicability of \ProposedMethodName{}.

\begin{table}[t]
  \centering
\begin{tabular}{lcc}
    \toprule
    \multirow{2}{*}{\textbf{Models}} & \multicolumn{2}{c}{\textbf{\ProposedBenchmark{}}}  \\
    \cmidrule(lr){2-3}
     &\textit{} Low Noise  & High Noise   \\
    \midrule
    TalkNet  \cite{tao2021someone} &  94.8 & 77.8\\
    \textbf{TalkNet + \ProposedMethodName{}} & \textbf{95.0} & \textbf{82.1} \\
    \midrule
     LoCoNet  \cite{wang2024loconet} & 96.2  & 86.7  \\
    \textbf{LoCoNet + \ProposedMethodName{}} &  \textbf{96.4} & \textbf{90.0} \\
    \bottomrule
    \end{tabular}
    \vspace{-1mm}
    \caption{\textbf{Results on \ProposedBenchmark} Combining \ProposedMethodName{} with TalkNet~\cite{tao2021someone} and LocoNet~\cite{tao2021someone} consistently improves performance on \ProposedBenchmark{}, with larger gains observed in scenarios with higher background noise.}
    \label{tab:noise}
    \vspace{-4mm}
\end{table}

\subsection{Evaluation on \ProposedBenchmark{}}

We further evaluate \ProposedMethodName{} on \ProposedBenchmark{}, treating it as an out-of-domain testbed for assessing robustness under real-world acoustic challenges. Models are trained on the AVA training set—composed primarily of clean, scripted movie scenes—and evaluated on the low-noise and high-noise subsets of \ProposedBenchmark{}, which feature modern online content with diverse and realistic background noise. This setup isolates the impact of acoustic variation without introducing major shifts in speaking style or visual context. 

As shown in Table~\ref{tab:noise}, LASER augmentation yields consistent gains: TalkNet’s mAP improves from 94.8 to 95.0 in low-noise conditions and from 77.8 to 82.1 in high-noise, a substantial improvement of 4.3 points. Similarly, LoCoNet improves from 96.2 to 96.4 in low-noise and from 86.7 to 90.0 in high-noise, gaining 3.3 points. Although both models are trained solely on AVA, \ProposedMethodName{} models generalize well to modern, noisy videos, with the largest gains observed under the most challenging acoustic conditions.

\begin{table*}[t]
    \centering
    \footnotesize
    
\begin{tabular}{ccc|ccc}
    \toprule
     &  &  & & \multicolumn{2}{c}{\textbf{\ProposedBenchmark{}}} \\
    \cmidrule(lr){5-6}
    \textbf{Trained with Lip Track Encoding (LTE)} & \textbf{Consistency Loss} & \textbf{Eval. with LTE} & \textbf{AVA-Val} & \textbf{Low Noise} & \textbf{High Noise} \\
    \midrule
    \xmark & \xmark & N/A   & 95.2 & 96.2 & 86.7 \\
    \cmark & \xmark & \cmark & 95.1 & 96.1 & 85.7 \\
    \cmark & \cmark & \cmark & 95.3 & 96.5 & 90.0 \\
    \cmark & \cmark & \xmark & 95.3 & 96.4 & 90.0 \\
    
    \bottomrule
\end{tabular}
\vspace{-2mm}
    \caption{\textbf{Importance of Each Component}. Training solely with Lip Track Encoding (LTE) leads to a slight decrease in mAP on both AVA-val~\cite{roth2020ava} and \ProposedBenchmark{}, primarily due to the unavailability of lips landmarks on some face tracks. The consistency loss addresses this limitation, enhancing the model's robustness.}
    \vspace{-2mm}
    \label{tab:main_ablation}
\end{table*}

\subsection{Evaluation with Unsynchronized Audios} 

\begin{table}[t]
  \centering

 \begin{tabular}{lccc}
    \toprule
    \multirow{1}{*}{\textbf{Model}} & \textit{} AVA & Talkies & ASW  \\
    \midrule
    LoCoNet \cite{wang2024loconet} & 75.9  & 72.1 & 56.7 \\
    \textbf{LoCoNet + \ProposedMethodName{}} &  \textbf{77.6} & \textbf{73.9} & \textbf{59.9} \\
    \bottomrule
  \end{tabular}
  \vspace{-1mm}
  \caption{\textbf{\textbf{Evaluation with swapped audio track}.} \ProposedMethodName{} significantly outperforms LoCoNet baseline when audio and visual frames are misaligned.}
  \vspace{-3mm}
  \label{tab:swap}
\end{table}

To test robustness against audio-visual misalignment, we simulate non-speaking scenarios by altering the audio stream while keeping the visual input unchanged. This reflects real-world situations like web conferencing, where off-screen or muted speakers may cause confusion. We consider two protocols on the AVA dataset: (1) audio-swapping, where the test audio is replaced with that from another video, and (2) audio-shifting, where a fixed delay is introduced. All experiments are conducted on the AVA test set, with accuracy computed as the average across all frames.

As shown in Figure~\ref{fig:unsync} and Table~\ref{tab:swap}, \ProposedMethodName{} consistently outperforms LoCoNet across both protocols. On audio-swapped AVA~\cite{roth2020ava}, Talkies~\cite{alcazar2021maas}, and ASW~\cite{kim2021look}, it improves accuracy by 1.7\%, 1.8\%, and 3.2\%, respectively. On audio-shifted AVA, it achieves at least a 2\% gain, and maintains strong performance across all temporal shifts on Talkies and ASW. These results highlight \ProposedMethodName{}’s robustness under unsynchronized audio conditions.


\subsection{Ablation Studies}
\label{sections:ablation_studies}

We next present ablation studies of \ProposedMethodName{}. All ablation studies are conducted with the LoCoNet framework~\cite{wang2024loconet} unless otherwise noted.

\subsubsection{Alternative Facial Landmark Integration}
\label{section:alternative}

\begin{table}[t]
  \centering

  \begin{tabular}{lcccc}
    \toprule
    \textbf{Model} & \textbf{AVA}& \textbf{Talkies} & \textbf{ASW} \\
    \midrule
    LoCoNet~\cite{wang2024loconet} & 95.2  & 88.4 & 88.4 \\
    ~w/ LP~\cite{Wang2019landmark_pooling} \textsuperscript{\S} & 94.7 & 88.3& 88.5\\
    ~w/ LDI~\cite{Geeroms2022_2d_encoding} \textsuperscript{\S} & 94.9 & 87.9 & 87.9 \\
    \textbf{~w/ \ProposedMethodName{}} &  \textbf{95.3} & \textbf{89.0} & \textbf{89.5} &\\
    \bottomrule
  \end{tabular}
  \vspace{-1mm}
  \caption{\textbf{Comparison of different ways to incorporate lip landmarks.} \textsuperscript{\S} Results are re-implemented in LoCoNet framework. \ProposedMethodName{} achieves the best performance over other alternative designs on both in-domain and out-of-domain evaluations.}
  \vspace{-5mm}
  \label{tab:landmark_comp}
\end{table}

Recall that we design a new encoding function to convert lip landmarks into 2D spatial features so that they can be integrated to visual features (Section~\ref{section:lte}). In this section, we compare our approach with other alternative designs to incorporate lip landmark information in to models.

\noindent \textbf{Landmark Pooling (LP).} Wang et al~\cite{Wang2019landmark_pooling} proposes to pool the visual features with lip landmark coordinates and use the resulting features in combination with original visual features for further processing. 

\noindent \textbf{Landmark as Discrete Inputs (LDI).} Geeroms et al.~\cite{Geeroms2022_2d_encoding} directly uses convolution layers to process the raw landmark coordinates. A lip landmark track $f \in \mathbb{R}^{T \times N \times 2}$, where $T$ denotes the number of frames, $N$ is the number of landmarks, and 2 is the coordinate $(x,y)$, is convolved with convolution layers along the temporal dimension. 

We implement these alternative designs in LoCoNet~\cite{wang2024loconet}  and compare them with our approach. As shown in Table~\ref{tab:landmark_comp}, \ProposedMethodName{} achieves the best performance over these baseline designs on both AVA and \ProposedBenchmark{}, highlighting its efficacy. Moreover, \ProposedMethodName{} only introduces negligible extra parameters to process lip landmarks compared to other approaches (details in Appendix).

\subsubsection{Importance of Each Component} 

The lip landmark encoding and consistency loss are central to \ProposedMethodName{}, enabling the integration of lip information during training while addressing missing lip landmarks at test time. In practice, our landmark detector fails on 15\% of AVA and 39\% of \ProposedBenchmark{} test face tracks. As shown in Table~\ref{tab:main_ablation}, using lip landmark encoding alone leads to a slight mAP drop due to missing test-time landmarks. In contrast, adding the consistency loss allows the model to retain strong performance without relying on lip landmarks, showing that the visual encoder learns to capture lip motion implicitly. This removes the need for lip landmarks during inference and reduces latency.

\begin{figure}[t!]
  \centering
    \includegraphics[width=\linewidth]{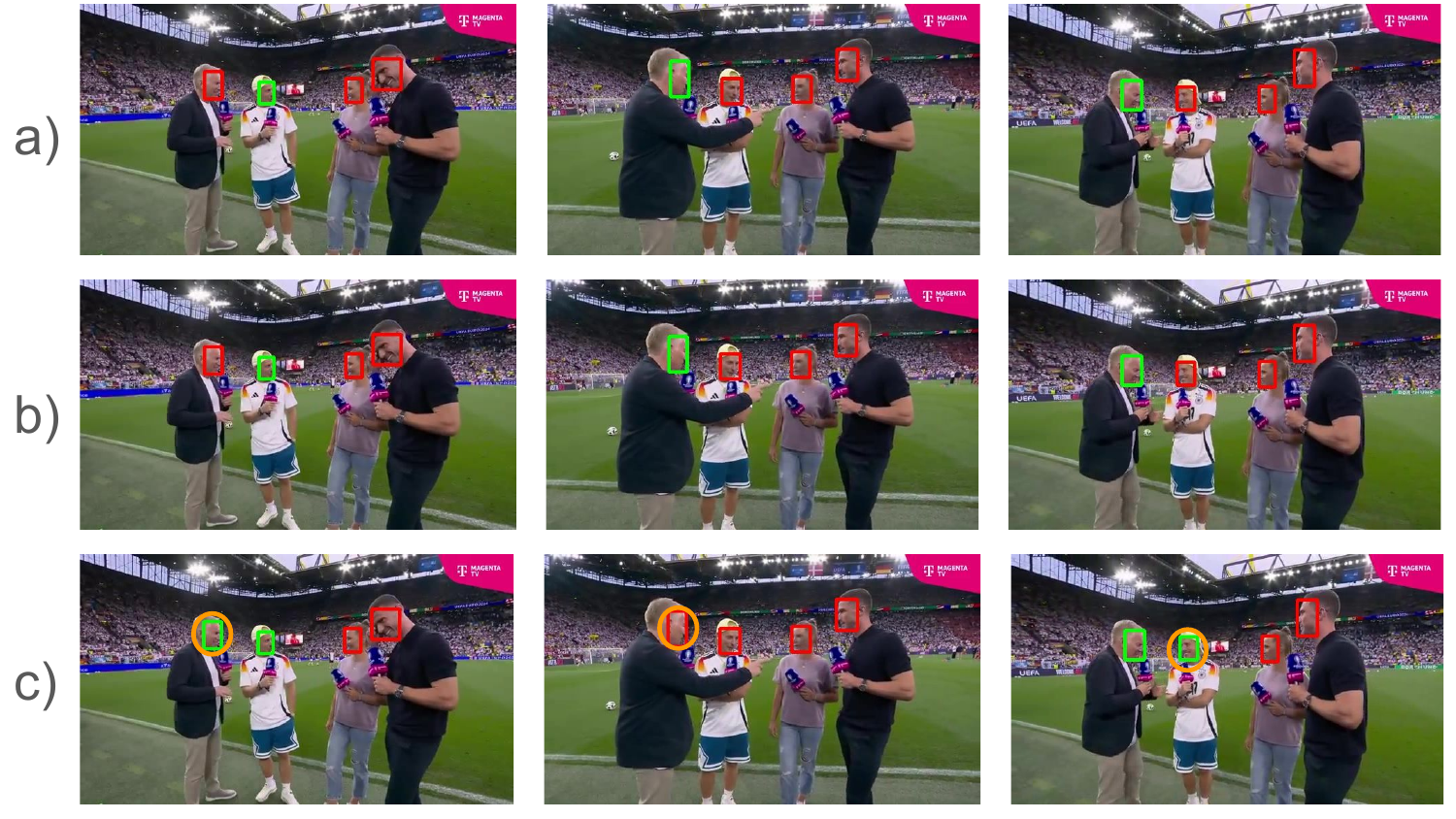}  
  \caption{\textbf{Qualitative Comparison on \ProposedBenchmark{}.} a) Ground-truth annotation. b) Detection results from our method \ProposedMethodName{}. c) Detection Results from baseline LoCoNet.  \textcolor{red}{Red}: not speaking; \textcolor{green}{Green}: speaking; \textcolor{orange}{orange}: incorrect predictions.}
  \vspace{-10pt}
  \label{fig:qualitative_result}
\end{figure}

\subsection{Qualitative Results}

As shown in Figure~\ref{fig:qualitative_result}, we present a qualitative comparison on LASER-Bench to illustrate how \ProposedMethodName{} handles high background noise (e.g., crowd cheering) more effectively than LoCoNet. In this clip, intense stadium noise overwhelms the audio signal, causing LoCoNet to miss genuine speaking events and produce false positives. In contrast, our method consistently localizes the true active speaker. This results in accurate speaking and non-speaking decisions where LoCoNet fails, highlighting the robustness of \ProposedMethodName{} in acoustically challenging, real-world scenarios.

\section{Conclusion}
\label{sections:conclusion}

In this work, we introduced \ProposedMethodName{} for active speaker detection. Unlike prior methods that rely on raw facial appearance, \ProposedMethodName{} explicitly guides the model to focus on lip landmarks, improving its ability to detect active speakers in challenging conditions. To bridge the gap in evaluating ASD performance under high background noise, we also introduced \ProposedBenchmark{} (LASER-Bench), a curated dataset featuring real-world videos with varying noise levels. Both quantitative and qualitative results show that \ProposedMethodName{} outperforms state-of-the-art models and achieves more reliable performance on LASER-Bench, demonstrating its robustness in noisy, unconstrained environments.
\section*{Acknowledgment}
This work was supported in part by NSF IIS2404180, and Institute of Information \& communications Technology Planning \& Evaluation (IITP) grants funded by the Korea government (MSIT) (No. 2022-0-00871, Development of AI Autonomy and Knowledge Enhancement for AI Agent Collaboration) and (No. RS-2022-00187238, Development of Large Korean Language Model Technology for Efficient Pre-training).
{
    \small
    \bibliographystyle{ieeenat_fullname}
    \bibliography{main}
}
\clearpage
\setcounter{figure}{0}
\setcounter{table}{0}
\setcounter{section}{0}
\renewcommand{\theequation}{\Alph{equation}}
\renewcommand{\thefigure}{\Alph{figure}}
\renewcommand{\thesection}{\Alph{section}}
\renewcommand{\thetable}{\Alph{table}}

\section{Appendix}
\label{sections:appendix_section}

\subsection{Training Details}
We implement our approach \ProposedMethodName{} in three different state-of-the-art ASD frameworks, LoCoNet~\cite{wang2024loconet}, TalkNet~\cite{tao2021someone}, and Light-ASD~\cite{liao2023light}. We closely follow the training details for fair comparison. Specifically, for LoCoNet-based training, we use batch size of 4 and sample 200 frames per training example. Each model is trained with 25 epochs on 4 RTX 2080 GPUs. For Light-ASD and TalkNet, we use a batch size that contains at most 2000 frames and each model is trained on one A40 GPU. We use Adam~\cite{diederik2014adam} as our optimizer with learning rate of $5 \times 10^{-5}$ that reduces by 0.5\% per epoch. For LoCoNet and TalkNet, we set $\lambda_{av} = 1$, $\lambda_a = 0.4$, and $\lambda_v = 0.4$; and for Light-ASD, we set $\lambda_{av} = 1$ and $\lambda_v = 0.5$, following the original work. Random resizing, cropping, horizontal flipping, and rotations are used as visual data augmentation operations and a randomly selected audio signal from the training set is added as background noise to the target audio~\cite{tao2021someone}.

\subsection{Implementation of Landmark Pooling and 2D Lip Encoding methods}

In this section, we introduce implementation details of the Landmark Pooling method~\cite{Wang2019landmark_pooling} and Landmark as Discrete Inputs~\cite{Geeroms2022_2d_encoding}. Since these methods were originally proposed to work with simple CNN architectures, we adapt them to the state-of-the-art LoCoNet framework for fair comparison.

\begin{figure*}[!t]
    \centering
    \includegraphics[width=\linewidth]{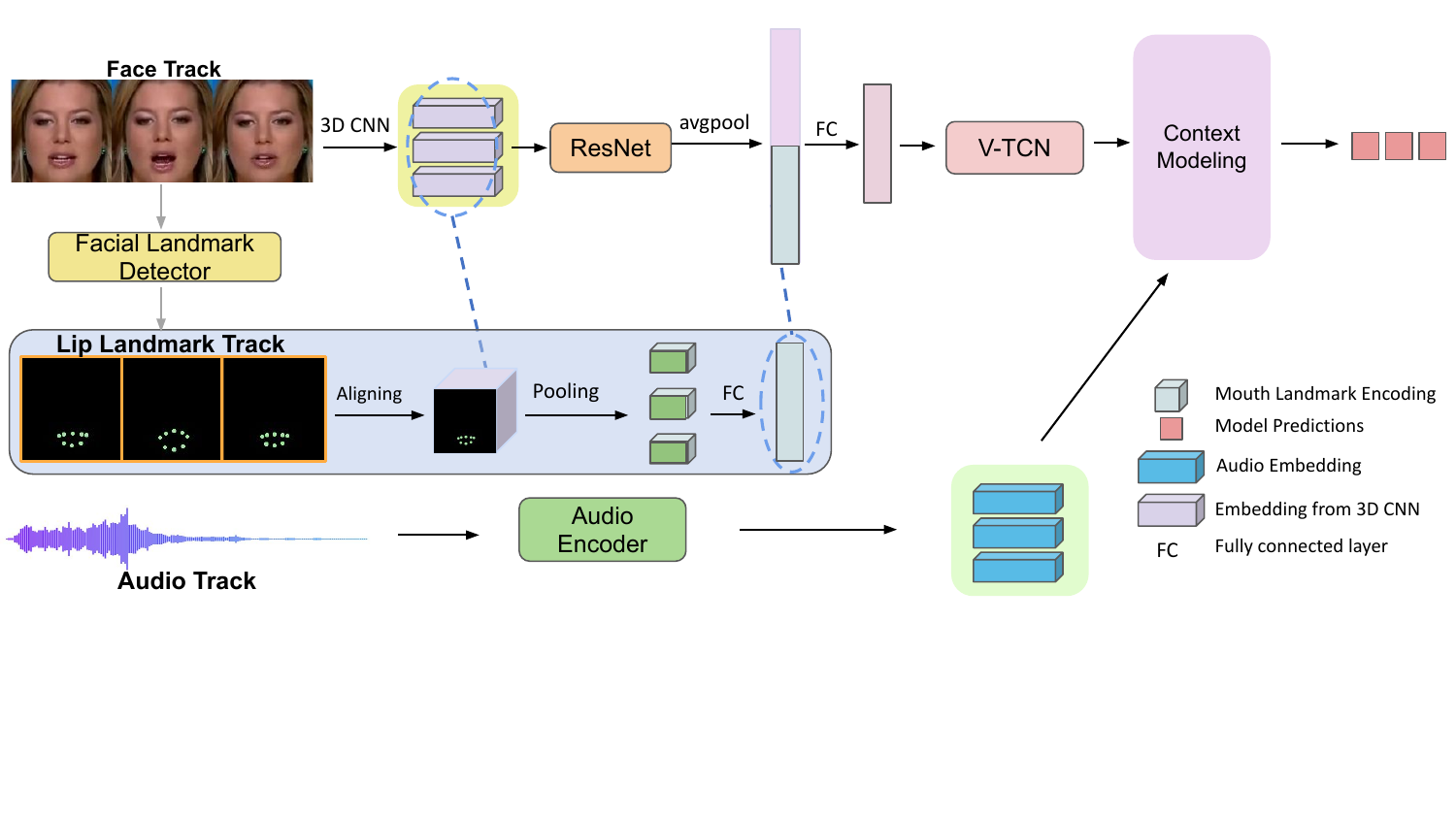}
    \vspace{-2.5cm}
    \caption{\textbf{Our implementation of Landmark Pooling \cite{Wang2019landmark_pooling}.} }
    \label{fig:landmark_pooling}
\end{figure*}

\subsubsection{Landmark Pooling}
Given a face track and the lips landmark detection results, we first obtain feature map $\mathcal{F}_v \in \mathbb{R}^{T \times C \times H \times W}$ after the first layer the ResNet, where $T$ denotes the number of frames, $H$ and $W$ are the spatial dimension of $f_v$, and $C$ is the number of channels, we follow~\cite{Wang2019landmark_pooling} to pool the feature with normalized lip landmark coordinates $\{(x_i, y_i)\}_{i=1}^N$ by $\mathcal{F}_v[x_i,y_i] \in \mathbb{R}^{T \times C}$. Then, we concatenate all $N$ landmarks into $\mathcal{F}_{ld} \in \mathbb{R}^{T \times NC}$ and feed $\mathcal{F}_{ld}$ through a fully connected layer to get $\mathcal{F}'_{ld} \in \mathbb{R}^{T \times D}$ where $D$ is the hidden dimension after the average pooling layer of ResNet. Finally, $\mathcal{F}'_{ld}$ is concatenated to $\mathcal{F}'_v$, the semantic feature output by ResNet and reduce the dimension with another fully-connected layer to obtain the final frame-level representation $\mathcal{F} \in \mathbb{R}^{T \times D}$. $\mathcal{F}$ is further processed by V-TCN and Context Modeling models along with audio feature to make prediction. The visualization of the method is presented in Figure~\ref{fig:landmark_pooling}.

\begin{figure*}[!t]
    \centering
    \includegraphics[width=\linewidth]{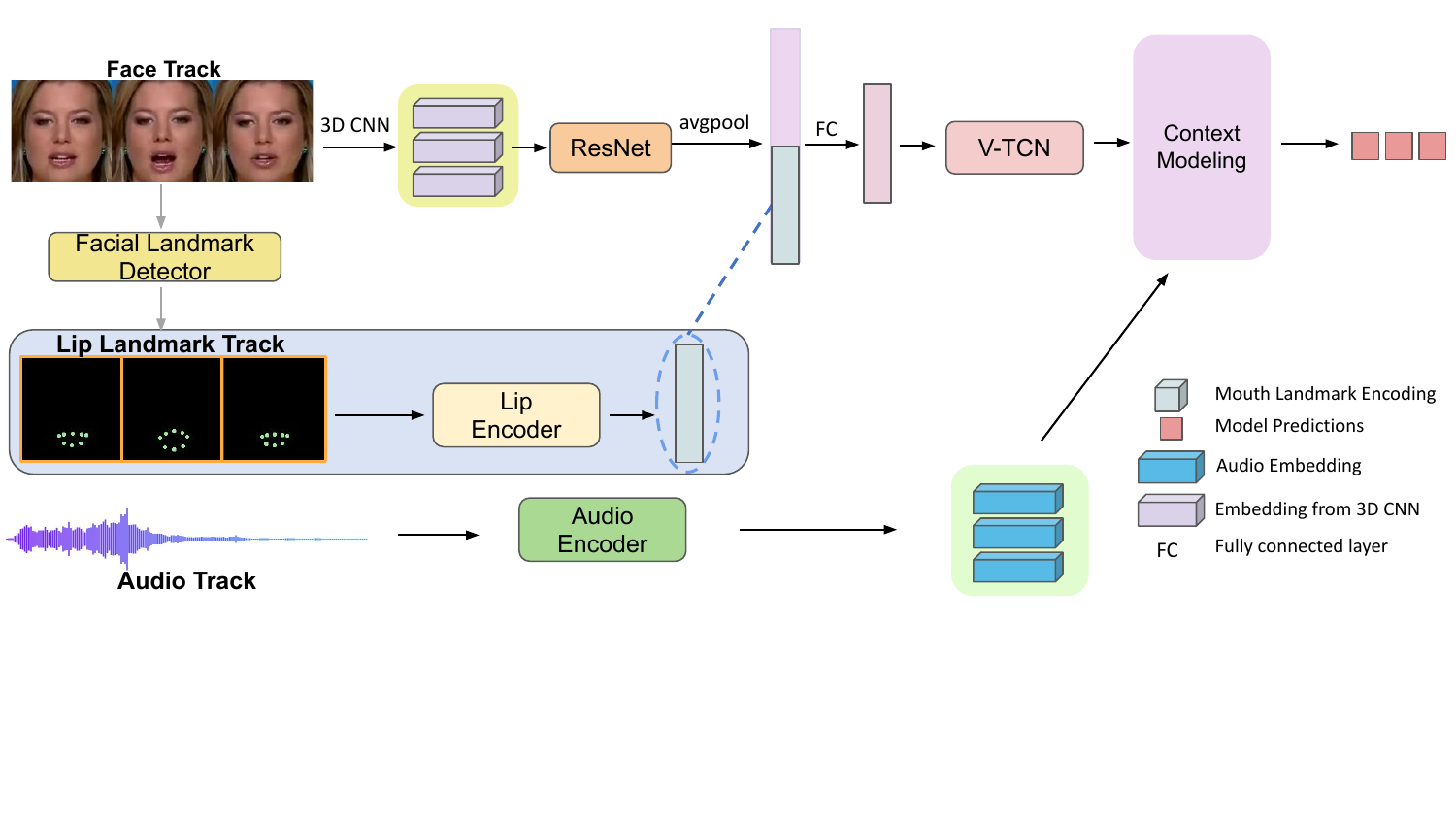}
    \vspace{-2.5cm}
    \caption{\textbf{Our implementation of 2D Lip Encoding method \cite{Geeroms2022_2d_encoding}.} }
    \label{fig:lip_encoding}
\end{figure*}

\subsubsection{Landmark as Discrete Inputs}
Given a face track and the lips landmark detection results $f \in \mathbb{R}^{T \times N \times 2}$ where $T$ is the number of frames, $N$ is the number of landmarks, and 2 is the coordinate $(x,y)$. We adopt the architecture in~\cite{Geeroms2022_2d_encoding} to process $f$ with 3x3 convolution layers and convert the raw landmark coordinates into embeddings. The resulting embedding is flattened and concatednated to the visual feature in the same way of Landmark Pooling. The visualization of the architecture is presented in Figure~\ref{fig:lip_encoding}.

\subsection{Parameter Effiency of \ProposedMethodName{}}
\begin{figure}
    \centering
    \includegraphics[width=1\linewidth]{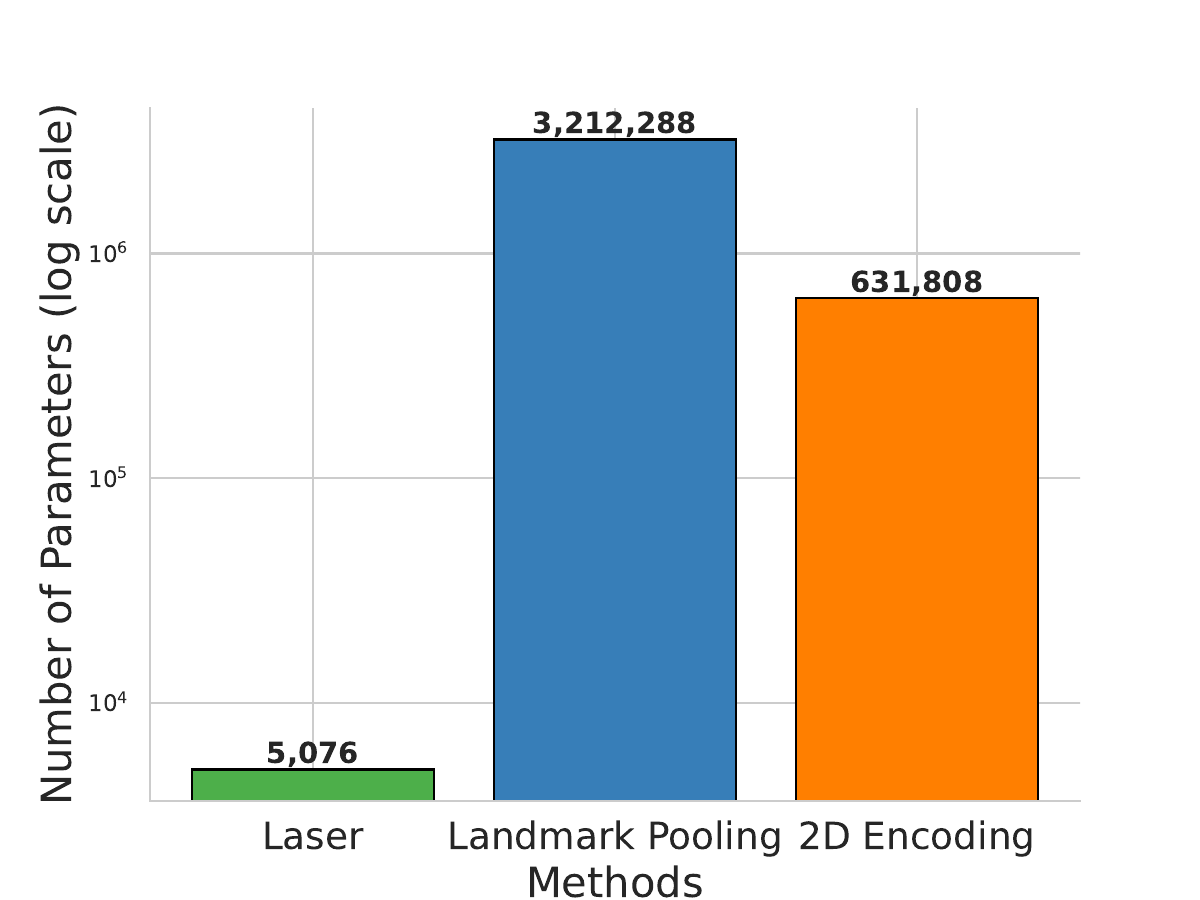}
    \caption{\textbf{Number of additional parameters of each method}. The y-axis is in log scale, and the number on each bar represents the true additional number of parameters. The x-axis contains 3 methods: LASER, Landmark Pooling \cite{Wang2019landmark_pooling}, and Landmark as Discrete Inputs \cite{Geeroms2022_2d_encoding}.}
    \label{fig:params}
\end{figure}

As illustrated in Figure~\ref{fig:params}, our proposed LASER approach not only achieves better performance but also only requires the negligible additional parameters compared to Landmark Pooling and Landmark as Discrete Inputs. This indicates that LASER’s design is significantly more parameter-efficient, which translates to reduced memory usage and faster training/inference. By minimizing the number of additional parameters without compromising accuracy, LASER offers a compelling balance between computational cost and performance gains.

\subsection{Hyperparameters Selection}
We present our ablation study on hyperparamete selection, including the ablation on consistency loss weight and the Lip Track Encoding injection layer. The hyperparameter selection process is conducted through a holdout evaluation protocol (different from all protocols in main evaluation) by reversing the audio in original video clip to establish a non-speaking scenario. This avoids the risk of tuning on the test set and ensures generalizability. 

\subsubsection{ResNet vs 3D CNN}
 As discussed in Section~\ref{sections:mler}, we integrate the aggregated lip track encoding (LTE) into the visual features produced by the 3D-CNN. We compare this approach with integrating LTE directly with the RGB face frames before the 3D-CNN. As shown in Table~\ref{tab:integration}, integrating LTE before the 3D-CNN leads to a significant performance drop. We speculate that this is due to the limited number of channels in the RGB image, where integrating LTE with additional channels may interfere with the feature extraction process of the original face frame. Thus, we opt to integrate LTE into the visual features output by the 3D-CNN.

 \subsubsection{Integrating Lip Track Encoding to Visual Features}

\begin{table}[h]
  \centering
\begin{tabular}{lccccc}
    \toprule
    \multirow{2}{*}{\textbf{Stage}} & \multicolumn{5}{c}{\textbf{\# Channels $S$}}  \\
    \cmidrule(lr){2-6}
    &\textit{} 1 & 2 & 4 & 8 & 16\\
    \midrule
    1 & 84.9 & 84.3 & \textbf{86.9} & 85.6 & 84.7\\
    2  & 84.3 & 84.7 & 86.2 & 83.6 & 85.0\\
    3  & 85.5 & 84.2 & 83.8 & 86.1 & 82.9\\
    4  & 85.3  & 86.5 & 85.1 & 84.9 & 82.4\\
    \bottomrule
    \end{tabular}
    \caption{\textbf{Ablation study on lip track integration with different stages of ResNet.} Setting $S=4$ and integrating lip track encoding before the first stage of ResNet results in the best performance.}
    \label{ablation:resnet}
\end{table}

\begin{table}[t!]
  \centering
\begin{tabular}{lcc}
    \toprule
    \textbf{Layer} & \textbf{Regular Audio} & \textbf{Noise Audio}\\
    \midrule
    3D CNN & 92.6 & 86.6\\
    ResNet  & \textbf{95.1} & \textbf{86.7}\\
    \bottomrule
    \end{tabular}
    \caption{\textbf{Ablation study on lip track encoding integration}. Integrating lip track encoding before 3D-CNN layer (i.e, on RGB frames) results in suboptimal performance.}
    \vspace{-10pt}
    \label{tab:integration}
\end{table}

We further investigate integrating lip landmark encoding (LTE) at different stages within ResNet, which has four internal stages with varying downsampling strides. Our ablation study (Table~\ref{ablation:resnet}) shows that integrating LTE before the first stage—using features from the 3D-CNN output—yields the best performance under noisy audio conditions, while later-stage integration degrades performance. Additionally, we examine the sensitivity of the aggregation factor $S$ when condensing sparse lip track encodings from $K=82$ to $S$ dense feature maps. We find that $S=4$ achieves the best balance, avoiding both excessive compression and sparsity.

\begin{table}[t!]
  \centering
\begin{tabular}{lccccc}
    \toprule
    \multirow{2}{*}{\textbf{Method}} & \multicolumn{5}{c}{\textbf{$\lambda_c$}}  \\
    \cmidrule(lr){2-6}
    & 0.2 & 0.4 & 0.6 & 0.8 & 1.0\\
    \midrule
    Acccuracy & 85.7 & 85.9 & 86.0 & 85.9 & \textbf{87.5}\\
    \bottomrule
    \end{tabular}
    \caption{\textbf{Ablation study on consistency loss weight.} Setting $\lambda_c = 1$ achieves the best results while decreasing $\lambda_c$ consistently degrades the performance.}
    \vspace{-15pt}
    \label{tab:weight}
\end{table}

\subsubsection{Weight of Consistency Loss} We also study the sensitivity of the weight of consistency loss \(\lambda_c\) where we range the value from 0.2 to 1.0. As shown in Table~\ref{tab:weight}, \(\lambda_c = 1\) achieves the best performance whereas decreasing the loss weight consistently hurts the model performance. Thus, we set \(\lambda_c = 1\) in all of our experiments.

\end{document}